\documentclass{esagnc}

\usepackage{bm}

\graphicspath{{figures/}}

\begin{document}

% Paper Title
\papertitle{On the Generation of a Synthetic Event-Based Vision Dataset for Navigation and Landing}

% Authors
\mainauthor{L. J. Azzalini}     % This goes in the footer
\author[(1)]{Loïc J. Azzalini}
\author[(2)]{Emmanuel Blazquez}
\author[(3)]{Gabriele Meoni}
\author[(4)]{Dario Izzo}
\author[(5)]{Alexander Hadjiivanov}
\affil[(1)]{\textit{Advanced Concepts Team, European Space Agency, European Space Research and Technology Centre (ESTEC), Keplerlaan 1, 2201 AZ Noordwijk, The Netherlands, jazzalin@outlook.com}}
\affil[(2)]{\textit{Advanced Concepts Team, European Space Agency, European Space Research and Technology Centre (ESTEC), Keplerlaan 1, 2201 AZ Noordwijk, The Netherlands}}
\affil[(3)]{\textit{Advanced Concepts Team, European Space Agency, European Space Research and Technology Centre (ESTEC), Keplerlaan 1, 2201 AZ Noordwijk, The Netherlands}}
\affil[(4)]{\textit{Advanced Concepts Team, European Space Agency, European Space Research and Technology Centre (ESTEC), Keplerlaan 1, 2201 AZ Noordwijk, The Netherlands, dario.izzo@esa.int}}
\affil[(5)]{\textit{Advanced Concepts Team, European Space Agency, European Space Research and Technology Centre (ESTEC), Keplerlaan 1, 2201 AZ Noordwijk, The Netherlands, alexander.hadjiivanov@esa.int}}

% Abstract
\paperabstract{
An event-based camera outputs an event whenever a change in scene brightness of a preset magnitude is detected at a particular pixel location in the sensor plane. The resulting sparse and asynchronous output, coupled with the high dynamic range and temporal resolution of this novel camera, motivate the study of event-based cameras for navigation and landing applications. However, the lack of real-world and synthetic datasets to support this line of research has limited its consideration for onboard use. This paper presents a methodology and a software pipeline for generating event-based vision datasets from optimal landing trajectories during the approach of a target body. We construct sequences of photorealistic images of the lunar surface with the Planet and Asteroid Natural Scene Generation Utility (PANGU) at different viewpoints along a set of optimal descent trajectories obtained by varying the boundary conditions. The generated image sequences are then converted into event streams by means of an event-based camera emulator. We demonstrate that the pipeline can generate realistic event-based representations of surface features by constructing a dataset of 500 trajectories, complete with event streams and motion field ground truth data. We anticipate that novel event-based vision datasets can be generated using this pipeline to support various spacecraft pose reconstruction problems given events as input, and we hope that the proposed methodology would attract the attention of researchers working at the intersection of neuromorphic vision and guidance navigation and control. Our code is available at \url{https://gitlab.com/EuropeanSpaceAgency/trajectory-to-events}.
%A larger dataset will serve as the base for an open competition on fully event-based landing hosted on the ACT competition platform.
}

% Keywords
\textbf{Keywords} --- Neuromorphic vision, event-based sensing, optical flow navigation, motion estimation

% Paper Content
\section{Introduction}

Neuromorphic vision technology holds great promise for space applications owing to its submillisecond temporal resolution, low power consumption and high dynamic range \cite{Roffe2021}\cite{Izzo2022}. Dynamic vision sensors, or event-based cameras as they are commonly known, have disrupted research fields across the computer vision landscape, from object tracking to image reconstruction, in robotics to the automotive industry \cite{Gallego2022}. A dynamic vision sensor is event-driven by design, as independently sensitive image pixels only respond to changes in scene brightness, leading to sparse and asynchronous output streams of data. Only motion between the scene and the camera will generate data, thus avoiding the capture of redundant static information and wasteful usage of onboard resources. Recent demonstrations of dynamic vision for in-orbit space situational awareness (SSA) \cite{Roffe2020}\cite{McHarg2022} are encouraging, warranting further adoption of event-based hardware in the space sector.

In recent years, there has been a surge in interest in event-based cameras for optimal control, landing and navigation. The most common method for translating events into control relies on estimating optical flow or divergence from events -- a bioinspired method used by bees and other insects for smooth landing on surfaces. Most studies on event-based landing consider the ventral landing case, without a lateral motion component. For instance, the use of optical flow reconstruction from events for optimal control of a micro air vehicle (MAV) during ventral landing was investigated in \cite{Hordijk2017}. The method relies on maintaining constant divergence (the ratio of the vertical velocity component to the height) to perform a smooth landing. A time-to-contact (TTC) metric derived from divergence estimated from event streams was used for optimal ventral landing in \cite{Sikorski2021}. Similarly, ventral landing scenarios based on a contrast maximisation technique for estimating divergence in a stream of events were studied in \cite{McLeod2023}. Recently, optical flow has been demonstrated for dynamic attitude, paving the way for navigation and control for small (insect-scale) robots without using accelerometers, as well as using micro-movements (e.g., vibrations) for attitude control and landing for larger vehicles \cite{deCroon2022_AccommodatingUnobservability}.

Whether tracking space debris or landing on a planetary body or asteroid, events are captured in the image plane of the event-based camera as a result of the relative dynamics between the onboard camera and the space environment. While event-based vision for guidance navigation and control (GNC) has been considered at a theoretical level in the past \cite{Izzo2022}, onboard use of this technology has yet to be demonstrated. Its limited adoption to date can, in part, be explained by the lack of event-based datasets tailored to vision-based navigation in space and state-of-the-art event processing tools found in other areas of application \cite{Afshar2020, Gehrig21}. Our main contribution is a software pipeline for generating event-based datasets from simulated spacecraft trajectories in a photorealistic scene generator for planetary bodies and asteroids.
% This work considers various landing scenarios (beyond the ventral case) in order to illustrate how the pipeline may be configured to capture event-based representation of surface features from various spacecraft poses.

We propose a flexible data pipeline which takes in trajectory specifications as input and outputs streams of events corresponding to the motion of surface features in the scene. From the initial and final conditions of the spacecraft trajectory and the properties of the target body, we solve an optimal control problem corresponding to a non-ventral, mass-optimal descent trajectory on a lunar-analogue surface. The trajectories are then used to manipulate the viewpoint of a pinhole camera model in the Planet and Asteroid Natural Scene Generation Utility (PANGU)\cite{Martin2021}, which renders synthetic images of the surface during approach. Finally, a video-to-event converter \cite{Hu2021, Gehrig2020, PROPHESEE} is used to generate synthetic events induced by the simulated landings. The sparse and asynchronous events include various sources of noise modelled after the performance of dynamic vision sensors in challenging lighting conditions in the field. The resulting dataset captures dynamic, event-based representations of common surface features such as craters and the Moon's horizon.
%Figure \ref{fig:pipeline} provides an overview of the pipeline, from inputs to outputs.

Section \ref{sec:background} introduces the dynamics of the lunar lander model and the mass-optimal control problem solved to simulate the landing trajectories. The section also reviews the idealised camera model and its equations of motion, leading to the derivation of motion field equations for planar and spherical landing surfaces. Finally, an overview of the principle of operation of the dynamic vision sensor is included to understand how an event-based camera may be emulated in software.
% describedSection \ref{sec:event_cam} introduces the alternative readout of the event-camera and how it can be emulated in software. As this work focuses on the generation of event streams to support navigation and landing, the processing of event-based data will not be discussed.
Section \ref{sec:pipeline} details the methodology that supports the generation of synthetic events from trajectory specifications using the emulated event-based camera. As the main contribution of this work, the pipeline is designed to be modular in order to support various spacecraft pose reconstruction problems given events as input. While the reconstruction of the dynamics from event-streams is not covered in this work, ground truth data is recorded in anticipation of such inverse problems. Aside from the optimal landing trajectories, the pipeline also saves motion field data in a format that is compatible with optical flow methods \cite{Baker2007} which have been successfully applied to event-streams in the past \cite{Benosman2014, Zhu18}.

A dataset of 500 trajectories that share characteristics with the typical braking, approach and descent stages of a lunar landing profile is generated. Highlights are presented in Section \ref{sec:results} to demonstrate the data generation capabilities of the pipeline. Specifically, the photorealistic scenes of a dimly illuminated lunar surface are compared to the corresponding motion fields and the synthetic event-based representations. The accumulation of events into frames reveal how the generation of events varies with various trajectory profiles and under different lighting conditions at the landing site. Finally, potential pipeline extensions, particularly to improve dataset variability, are discussed in Section \ref{sec:discussion} as follow-up work.

\section{Theoretical Background}\label{sec:background}

\begin{figure}[ht]
  \centering
  \includegraphics[width=\linewidth]{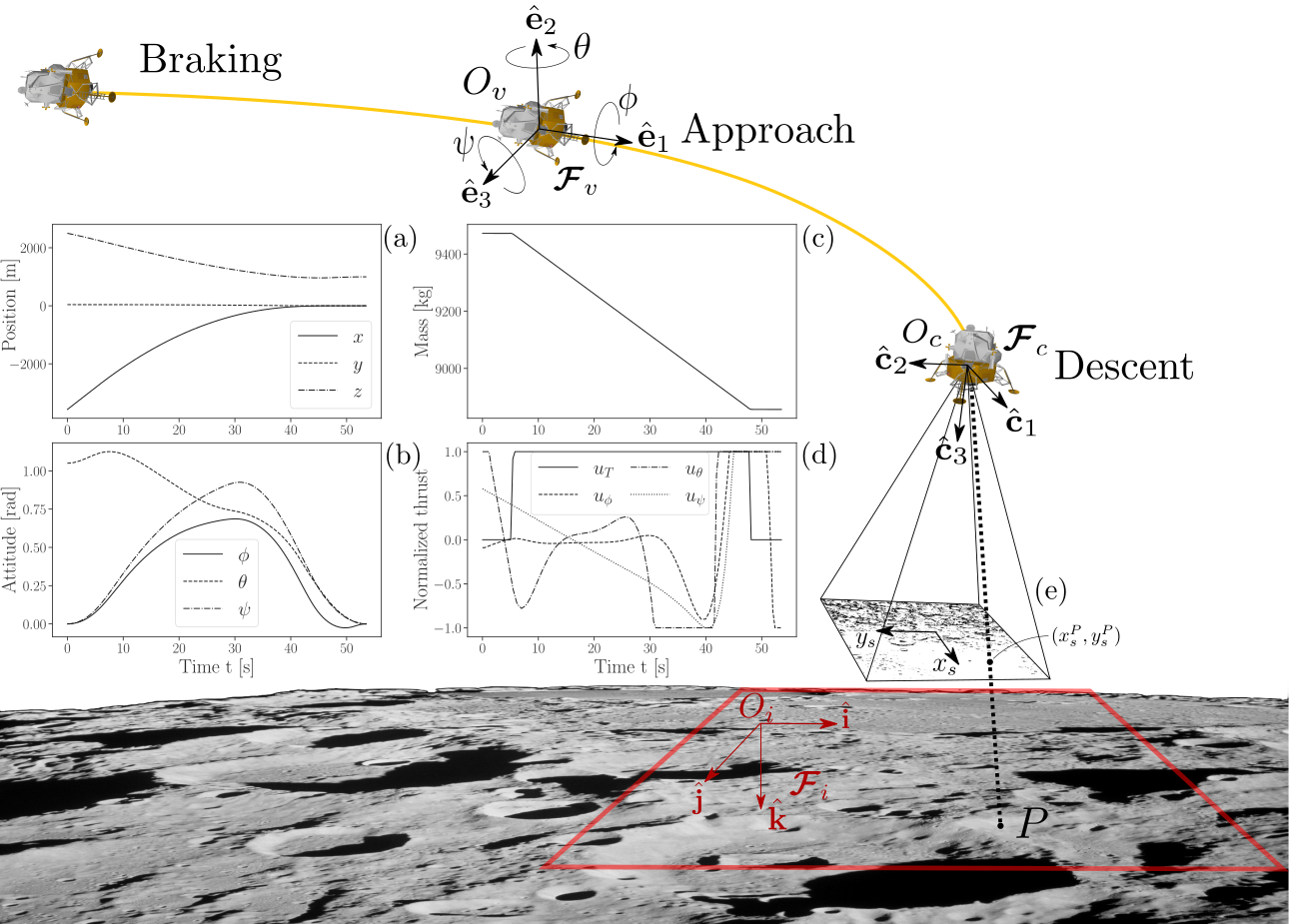}
  \caption[]{\small Geometry of the landing scenario and sample optimal trajectory: (a) position, (b) attitude, (c) mass and (d) normalised thrusts of the spacecraft as a function of time; (e) projection of the 3D point $P$ into the 2D image plane position $(x_s^P, y_s^P)$ for an idealised pinhole descent camera}
  \label{fig:lunar_landing_profile}
\end{figure}

We consider the problem of a lunar lander entering the final stages of a descent towards the surface. In this work, it is assumed that an event-based descent camera is mounted underneath the lunar module and pointed in the direction of motion, offering uninterrupted streams of data to the onboard computer to support vision-based navigation. The geometry of the problem is illustrated in Figure \ref{fig:lunar_landing_profile} with a simplified trajectory broken down into a braking, approach and (terminal) descent phase, and an idealised camera model capturing surface features in its image plane.
% TODO: add letters to figure to detail in the caption

\subsection{Equations of motion}\label{sec:eom}
The inertial frame of reference $\bm{\mathcal{F}}_i = \begin{bmatrix}\hat{\mathbf{i}}&\hat{\mathbf{j}}&\hat{\mathbf{k}}\end{bmatrix}^T$ is placed on the lunar surface at the location of the landing site $O_i$. The attitude of the spacecraft with body frame $\bm{\mathcal{F}}_{v}$ relative to the inertial frame is characterised by the roll-pitch-yaw sequence of Euler angles $(\theta_1, \theta_2, \theta_3) = (\phi, \theta, \psi)$. The camera frame $\bm{\mathcal{F}}_c$ and body frame $\bm{\mathcal{F}}_{v}$ are made to be coincident, such that the camera centre $O_c$ (centre of projection) is identical to $O_{v}$. The position, the velocity and angular velocity of the vehicle with respect to the inertial frame are respectively defined as
\begin{align}
\mathbf{r}_{i}^{vi} &= \bm{\mathcal{F}}_{i}^T \left[
\begin{array}{l}
x\\y\\z
\end{array}
\right] = \bm{\mathcal{F}}_{i}^T \mathbf{r},&
\mathbf{v}_{i}^{vi} &= \bm{\mathcal{F}}_{i}^T \left[
\begin{array}{l}
v_x\\v_y\\v_z
\end{array}
\right] = \bm{\mathcal{F}}_{i}^T \mathbf{v},&
\boldsymbol{\omega}_{vi} &= \bm{\mathcal{F}}_{v}^T \left[
\begin{array}{l}
p\\q\\r
\end{array}
\right] = \bm{\mathcal{F}}_{v}^T \boldsymbol{\omega} 
\end{align}
and the transformation from $\bm{\mathcal{F}}_{v}$ to $\bm{\mathcal{F}}_{i}$, i.e. $\mathbf{R}_{iv} = \bm{\mathcal{F}}_{i}\cdot\bm{\mathcal{F}}_{v}^T$, is described by the direction cosine matrix
\begin{equation}\label{eq:dcm}
    \mathbf{R}_{iv} =
    \begin{bmatrix}
        \cos{\theta}\cos{\psi} & -\cos{\phi}\sin{\psi}+\sin{\phi}\sin{\theta}\cos{\psi} & \sin{\phi}\sin{\psi}+\cos{\phi}\sin{\theta}\cos{\psi} \\
        \cos{\theta}\sin{\psi} & \cos{\phi}\cos{\psi}+\sin{\phi}\sin{\theta}\sin{\psi} & -\sin{\phi}\cos{\psi}+\cos{\phi}\sin{\theta}\sin{\psi} \\
        -\sin{\theta} & \sin{\phi}\cos{\theta} & \cos{\phi}\cos{\theta}
    \end{bmatrix} = \mathbf{R}
\end{equation}
The throttle $u_T\in [0,1]$ regulates the spacecraft main engine (normalised with respect to maximum primary thrust $\bar{F}_a$), while the throttles $u_\phi, u_\theta, u_\psi \in [-1,1]$ regulate the attitude control nozzles (normalised with respect to maximum secondary thrust $\bar{F}_b$). These are placed symmetrically at a distance $L$ from the spacecraft centre of mass to generate roll, pitch and yaw moments. Then, we consider the following 12-DOF lander dynamics:
\begin{align}
\dot{\mathbf{r}} & = \mathbf{v} \label{eq:eom_pos}\\
\dot{\mathbf{v}} &= 
    \begin{bmatrix}
        0 \\
        0 \\
        g
    \end{bmatrix} - \frac{1}{m} \mathbf{R}
    \begin{bmatrix}
        0 \\
        0 \\
        \bar{F}_a  u_T
    \end{bmatrix}\label{eq:eom_vel}\\
    \begin{bmatrix}
    \dot{\phi} \\
    \dot{\theta} \\
    \dot{\psi}
    \end{bmatrix}
     &=
    \begin{bmatrix}
    1 & \sin{\phi} \tan{\theta}     & \cos{\phi} \tan{\theta} \\
    0 & \cos{\phi}                  & -\sin{\phi} \\
    0 & \sin{\phi} / \cos{\theta}   & \cos{\phi} / \cos{\theta}
    \end{bmatrix}
    \begin{bmatrix}
        p \\
        q \\
        r\\
    \end{bmatrix} \label{eq:eom_att}\\
    \mathbf{I} \dot{\boldsymbol{\omega}} &= \boldsymbol{\omega} \times \boldsymbol{I} \boldsymbol{\omega} +  2 L \bar{F}_b \begin{bmatrix}
     u_\phi \\
     u_\theta \\
     u_\psi \\
\end{bmatrix}\label{eq:eom_pqr}
\end{align}
and the mass-thrust relationship
\begin{equation}\label{eq:mass_roc}
    \dot m = -\frac{\overline{F}_a u_T + 2 \overline{F}_b
    (u_\phi+u_\theta+u_\psi)}{I_{sp}g_0}    
\end{equation}
where explicit time dependence notation has been dropped to improve readability. The gravitational accelerations on the landing body and Earth are respectively given by $g$ and $g_0$, and $I_{sp}$ is the specific impulse of the propulsion system. We let $\mathbf{x}(t) = \begin{bmatrix}\mathbf{r}(t)^T&\mathbf{v}(t)^T&\phi(t)&\theta(t)&\psi(t)&\boldsymbol\omega(t)^T&m(t)\end{bmatrix}^T$ and $\mathbf{u}(t) = \begin{bmatrix}u_T(t) & u_{\phi}(t) & u_{\theta}(t) & u_{\psi}(t)\end{bmatrix}^T$ be the states and inputs vectors such that the dynamics may be concisely written as $\dot{\mathbf{x}}(t) = \mathbf{f}(\mathbf{x}(t), \mathbf{u}(t))$.

\subsection{Optimal Control Problem}\label{sec:ocp}

The dataset trajectory generation is based on a mass-optimal control problem (OCP) that follows from Pontryagin's maximum principle. The objective is to maximise the final lander mass $m(t_f)$ while minimising the total energy expended to bring the spacecraft from a starting pose to a final pose, subject to vehicle dynamics and thrust limits. Formally, the problem consists in minimising
\begin{equation}\label{eq:ocp_cost}
    J(m(t), \mathbf{u}(t)) = -(1-\epsilon) \int_{0}^{t_f} \dot{m}(\tau) \, d\tau + \epsilon \int_{0}^{t_f} \mathbf{u}(\tau)^T\mathbf{u}(\tau) \, d\tau
\end{equation}
subject to
\begin{equation}
\begin{aligned}\label{eq:ocp_constraints}
    &\dot{\mathbf{x}}(t) = \mathbf{f}(\mathbf{x}(t), \mathbf{u}(t)), && \qquad\text{Dynamic constraints} \\
    &\cos\phi \cdot \cos\theta \geq \cos\lambda, && \qquad\text{Max thrust tilt} \\
    &-\mathbf{1} \leq \mathbf{u}(t) \leq \mathbf{1}, && \qquad\text{Thrust limits} \\
    &\mathbf{x}(0) = \mathbf{x}_0,\;\mathbf{x}(t_f) = \mathbf{x}_f && \qquad\text{Boundary conditions}
\end{aligned}
\end{equation}
where $\epsilon > 0$ weighs the relative importance of the final mass and energy expenditure, and $\lambda$ denotes the maximum allowable tilt angle. The initial states $\mathbf{x}_0$ and final states $\mathbf{x}_f$ are to be specified in a manner that approximates the desired braking, approach or descent trajectory characteristics in a free or pinpoint landing. Figure \ref{fig:lunar_landing_profile} illustrates the solution of an OCP which has been parameterised to simulate a mass-optimal approach phase, i.e. the high starting pitch is regulated in preparation of the terminal descent.

\subsection{Camera Model and Motion Field}\label{sec:cam_eom}

Before considering the event-based representation of visual features, we review the relationship between a 3D point in the scene and its projection in the image plane. As illustrated in Figure \ref{fig:lunar_landing_profile}, we consider the generic point $P$ on the planetary surface and let
\begin{equation}
% \vett{O_cP} = \bm{\mathcal{F}}_{c}^T \begin{bmatrix}X\\Y\\Z\end{bmatrix}
\mathbf{r}_c^{O_c P} = \bm{\mathcal{F}}_{c}^T \begin{bmatrix}X\\Y\\Z\end{bmatrix} = \mathbf{r}_v^{O_c P}
\end{equation}
As the spacecraft moves relative to the surface, the apparent velocity of the point $P$ in the camera frame $\bm{\mathcal{F}}_c$ is opposite to its velocity in $\bm{\mathcal{F}}_i$:
\begin{equation}\label{eq:motion_p}
% \dot{\mathbf{r}}_c^{O_c P} = \bm{\mathcal{F}}_{c}^T \begin{bmatrix}\dot X\\\dot Y\\\dot Z\end{bmatrix} = -\mathbf{v}_{O_c} - \boldsymbol\omega \times \vett{O_cP}
\dot{\mathbf{r}}_c^{O_c P} = \bm{\mathcal{F}}_{c}^T \begin{bmatrix}\dot X\\\dot Y\\\dot Z\end{bmatrix} = -\mathbf{R}_{iv}^T \mathbf{v}_{i}^{vi} - \boldsymbol\omega_{vi} \times \mathbf{r}_{v}^{O_c P}
\end{equation}
with $\mathbf{v}_{c}^{vi} = \mathbf{R}_{iv}^T \mathbf{v}_{i}^{vi}$ being the inertial velocity of $O_c$ in the camera frame since $\bm{\mathcal{F}}_c$ and $\bm{\mathcal{F}}_{v}$ are coincident. Letting $\mathbf{v}_{c}^{vi} = \begin{bmatrix}v_{c,x} & v_{c,y} & v_{c,z}\end{bmatrix}^T$ and expanding \eqref{eq:motion_p} yields
\begin{equation}\label{eq:vel_p}
\begin{bmatrix}
\dot X\\\dot Y\\\dot Z
\end{bmatrix} =
-\begin{bmatrix}
v_{c,x}\\v_{c,y}\\v_{c,z}
\end{bmatrix}
- \begin{bmatrix}
qZ-rY\\rX-pZ\\pY-qX
\end{bmatrix}
\end{equation}
% TODO: provide better explanation of projection of 3D world point into image coordinates; describe the camera intrinsics parameters K here instead of in the Results section
% Then, given an idealised pinhole camera with focal length $f$, equal aspect ratio and no skew, the projection of $P$ in the sensor array corresponds to
% \begin{align}
% \begin{split}
%     \begin{bmatrix}
%         w x_s \\ w y_s \\ w
%     \end{bmatrix} =
%     \mathbf{K}
%     \begin{bmatrix}
%      \mathbf{I} | \mathbf{0}
%     \end{bmatrix}
%     \begin{bmatrix}
%         X \\ Y \\ Z \\ 1
%     \end{bmatrix} = 
%     % \begin{bmatrix}
%     %     w x_s \\ w y_s \\ w
%     % \end{bmatrix} = 
%     \begin{bmatrix}
%         f & 0 & c_x & 0 \\
%         0 & f & c_y & 0 \\
%         0 & 0 & 1   & 0
%     \end{bmatrix}
%     \begin{bmatrix}
%         X \\ Y \\ Z \\ 1
%     \end{bmatrix}
% \end{split}
% \end{align}
Then, given an idealised pinhole camera with focal length $f$, the projection of $P$ in the sensor array corresponds to
% such that
\begin{equation}\label{eq:proj}
    x_s = f\frac{X}{Z},\qquad y_s = f \frac{Y}{Z}
\end{equation}
By differentiating \eqref{eq:proj} and substituting in \eqref{eq:vel_p}, we obtain
\begin{align}\label{eq:mf}
    \begin{bmatrix}
        u \\ v
    \end{bmatrix} &= \frac{1}{Z}
    \begin{bmatrix}
    -f & 0 & x_s \\
    0 & -f & y_s
    \end{bmatrix}
    \begin{bmatrix}
        v_{c,x} \\ v_{c,y} \\ v_{c,z}
    \end{bmatrix} +
    \begin{bmatrix}
    \frac{x_s y_s}{f} & -(f + \frac{x_s^2}{f}) & y_s \\
    f + \frac{y_s^2}{f} & - \frac{x_s y_s}{f} & -x_s
    \end{bmatrix}
    \begin{bmatrix}
        p \\
        q \\
        r
    \end{bmatrix}
\end{align}
where $u$ and $v$ describe the motion field induced in the image plane as a result of the dynamics of the spacecraft relative to the scene (i.e., ego-motion model). 
The depth $Z$ of a point can be substituted in, provided appropriate complementary sources of information about the target body (e.g. digital elevation models). Given limited \textit{a priori} information, we can alternatively substitute the inverse depth map $h(x_s, y_s) = \frac{1}{Z}$ for an expression that only depends on the spacecraft state by assuming the local shape of the target landing site; planar and spherical geometries are considered below. We note that one of the inverse problems mentioned above consists in estimating $h(x_s, y_s)$ (up to an arbitrary scale) by inverting \eqref{eq:mf} as discussed in Section \ref{sec:pipeline_mf}.

\subsubsection{Planar landing site}

If the landing site is assumed to be a perfect plane, the point on the surface $P$ relates to the spacecraft altitude as follows:
\begin{equation}
\begin{split}
        \hat{\mathbf{k}}_{c} \cdot \mathbf{r}_c^{O_c P} &= \alpha X + \beta Y + \gamma Z = -z = H\\
        Z &= \frac{H}{\alpha x_s + \beta y_s + \gamma} = \frac{H}{A}
\end{split}
\end{equation}
where $\hat{\mathbf{k}}_c = \bm{\mathcal{F}}_c^T \begin{bmatrix}\alpha&\beta&\gamma\end{bmatrix}^T$ and the altitude $H$ is equivalent to the negative height from the surface (see the problem geometry in Figure \ref{fig:lunar_landing_profile}). Given that $\bm{\mathcal{F}}_c$ and $\bm{\mathcal{F}}_{v}$ are taken to be coincident in this study, we have from \eqref{eq:dcm}
\begin{equation}
    \alpha = -\sin\theta,\qquad \beta = \sin\phi\cos\theta,\qquad \gamma = \cos\phi\cos\theta
\end{equation}
and hence, letting $A = -x_s \sin\theta + y_s \sin\phi\cos\theta + \cos\phi\cos\theta$,
\begin{align}
    h(x_s, y_s) &= \frac{A}{H}
\end{align}

\subsubsection{Spherical landing site}\label{sec:mf_sphere}

Alternatively, if the descent camera points towards a perfect spherical surface of radius $R$ centered at $O_R$, the surface point $P$ relates to the spacecraft altitude $H$ as follows:
\begin{align}
\begin{split}
    % \vett{O_R P} &= \vett{O_R O_c}  + \vett{O_cP} \\
    % &= -(R + H)\hat{\mathbf{k}} + \vett{O_cP}
    \mathbf{r}_{c}^{O_R P} &= \mathbf{r}_{c}^{O_R O_c}  + \mathbf{r}_{c}^{O_c P} \\
    &= -(R + H)\hat{\mathbf{k}}_c + \mathbf{r}_{c}^{O_c P}
\end{split}
\end{align}
Given that $(\mathbf{r}_{c}^{O_c P} - (R+H) \hat{\mathbf{k}}_c) \cdot (\mathbf{r}_{c}^{O_c P} - (R+H) \hat{\mathbf{k}}_c) = R^2$,
\begin{align}
\begin{split}
    (X-\alpha(R+H))^2 + (Y-\beta(R+H))^2 + (Z-\gamma(R+H))^2 &= R^2 \\
    X^2+Y^2+Z^2-2(R+H)(\alpha X+ \beta Y+ \gamma Z)+(R+H)^2 - R^2 &= 0
\end{split}
\end{align}
which, after rearranging \eqref{eq:proj} and substituting for $X$ and $Y$, transforms into
\begin{equation}
    (x_s^2+y_s^2+1)Z^2-2(R+H)(\alpha x_s+ \beta y_s+ \gamma )Z+H^2+2RH = 0
\end{equation}
Solving the quadratic equation in $Z$ and inverting yields the inverse depth map
\begin{equation}
    h(x_s, y_s) = \frac{x_s^2 + y_s^2 + 1}{(R+H)A - \sqrt{(R+H)^2 A - 2RH(x_s^2 + y_s^2 + 1)}}
\end{equation}

\subsection{Event-Based Camera Emulation}\label{sec:event_cam}

\begin{figure}[ht]
  \centering
  \includegraphics[width=\linewidth]{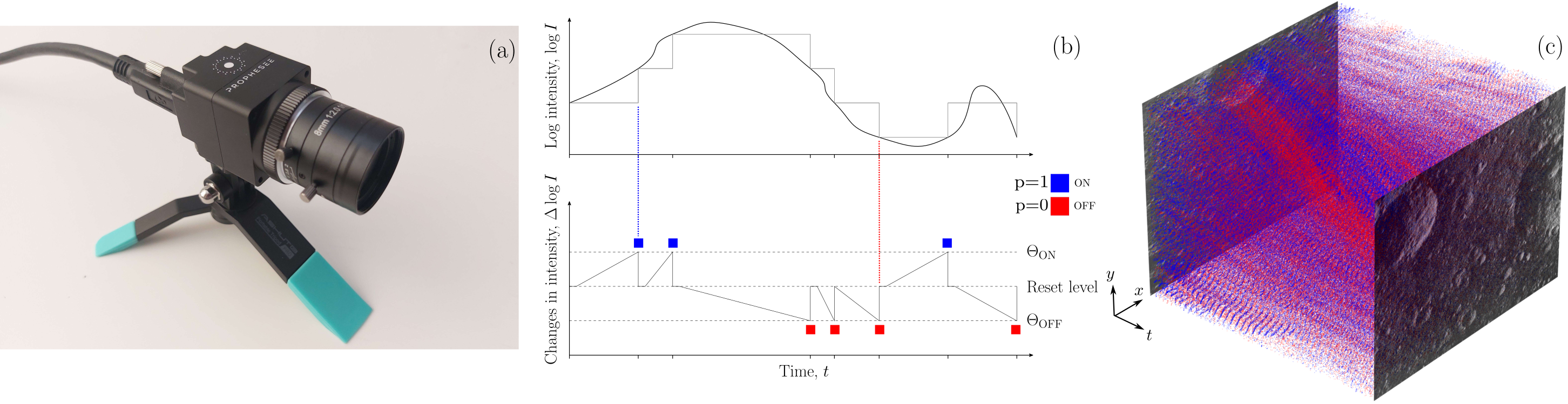}
  \caption{Event-based camera: (a) the EVK-4 HD event-based vision sensor \cite{PROPHESEE}, (b) the operation principle of one of its pixels (c) and the spatiotemporal representation of synthetic event-frames between two intensity frames capturing a descent with significant yaw motion}
  \label{fig:event_camera}
\end{figure}

Having described the geometry of an idealised pinhole camera and image sensor, we now introduce an alternative to the standard frame-based readout. While a standard frame camera outputs intensity frames at a fixed rate based on an external clock signal, the pixels of an event-camera asynchronously report changes in scene brightness (log intensity, $\log I$), resulting in a stream of discrete \textit{events}. Figure \ref{fig:event_camera} (a) shows an example of event-based vision hardware, whereas (c) schematically represents the spatiotemporal stream of raw events that could be obtained during a landing scenario. 

The operation principle of a dynamic vision pixel is differential in nature: rather than accumulating light for a certain amount of exposure time, like traditional frame-based sensors, they register changes in illumination (i.e., $\Delta \log I$). Since each pixel can detect both positive and negative changes in brightness, there are separate thresholds $\Theta_{ON}$ and $\Theta_{OFF}$ for positive and negative events, as depicted in Figure \ref{fig:event_camera} (b) in blue and red respectively. 
Once the brightness changes beyond one of the thresholds, the pixel sets its new baseline to the current brightness level (i.e., memory) and outputs an event with positive (negative) polarity indicating an increase (decrease) in perceived brightness. The readout of a single event takes the form of $(t, x, y, p)$, where $t$ is the timestamp of the event (typically in the order of microseconds), $p$ indicates its polarity (positive or negative), and $(x, y)$ represents the pixel coordinates. The lower data generation rate compared to a frame sensor allows for readout rates in the range of $2-1200\,\rm{MHz}$, submillisecond temporal resolution and low power consumption \cite{Gallego2022}, whereas the sensitivity and logarithmic scale of the independent pixels afford the sensor a very high dynamic range. As an example, Table \ref{tab:evk4} lists the corresponding specifications for the event-based camera shown in Figure \ref{fig:event_camera} (a).
\begin{table}[ht]
\begin{center}
\caption{Example specifications of event-based vision hardware (EVK-4 HD \cite{PROPHESEE})}\label{tab:evk4}
\begin{tabular}{@{}lcc@{}}
\toprule
Property & Range & Units \\
\midrule
Resolution & $1280\times 720$ & $\rm{px}$\\
Dynamic range & $>120$ & $\rm{dB}$\\
Latency & $220$ & $\rm{\mu s}$\\
Power consumption & $0.5$ & $\rm{W}$\\
\bottomrule
\end{tabular}
\end{center}
\end{table}

In order to consider an event-based camera (a data-driven sensor), for navigation and landing, we need to resort to hardware emulation given the difficulty in sourcing such data \textit{in situ}. Fortunately, video-to-event toolboxes such as the one provided by \cite{PROPHESEE} (based on \cite{Gehrig2020, Hu2021}) have been designed to faithfully replicate the output of event cameras in software, including common sources of noise based on the performance of the sensors in various lighting conditions \cite{Hu2021}. The video-to-event pipeline consists of several processing steps designed to convert frames into event sequences or streams. First, the frames of the original RGB video are converted to grayscale based on the brightness $Y$ of the scene at each pixel; the converted frames are assigned timestamps derived from the frame rate. The next step involves converting the linear frame brightness scale to the logarithmic event brightness scale. This is particularly important due to the limited dynamic range of computer graphics formats (the brightness information content is usually limited to 8 bits, meaning that the brightness is limited to values between $0$ and $255$). Therefore, the video-to-event converter linearises the logarithmic scale for low-intensity values ($Y < 20$) in order to reduce quantisation noise in the output. As the final preprocessing step before event generation, the output of the log-linear mapping step is fed into an infinite impulse response (IIR) low-pass filter that models the physical response profile of an analog pixel in low-light conditions, where the bandwidth of the filter increases proportionally to the brightness value $Y$.

The video-to-event pipeline also models effects observed in real hardware pixels. In particular, it emulates hot pixels that produce events at a high rate even in the absence of input, leak current resulting in sporadic ON events (typically with a rate of about $0.1 \rm Hz$) and shot noise observed due to the disproportionate contribution of individual photons to the integration of photon counts in very dark conditions (modelled as a Poisson process). Figure \ref{fig:event_camera} (c) shows how features in the scene give rise to spiralling spatiotemporal streams of postive/negative events as a result of a descent towards the lunar surface with the emulated event-camera pointing straight down (see Figure \ref{fig:lunar_landing_profile}).

\section{Trajectory-to-Event Pipeline}\label{sec:pipeline}

\begin{figure}[ht]
  \centering
  \includegraphics[width=\linewidth]{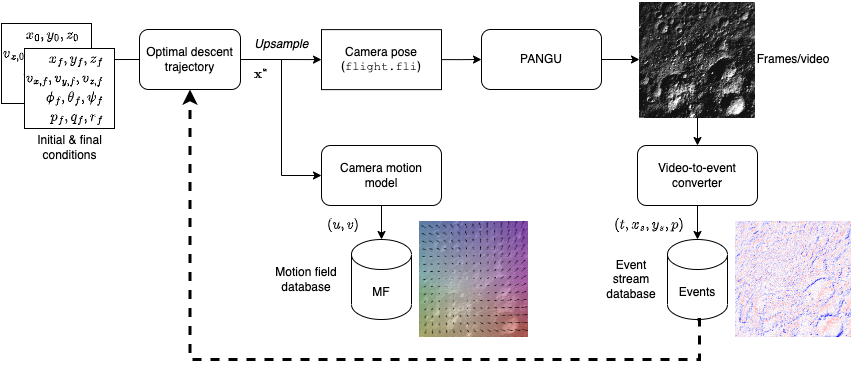}
  \caption{\small Pipeline to generate event-based vision datasets: input trajectory specifications are processed into ground truth motion fields, video simulations of the landing sequence and streams of asynchronous events}
  \label{fig:pipeline}
\end{figure}

The different modules of the trajectory-to-event pipeline are depicted in Figure \ref{fig:pipeline}. Given initial and final conditions for the lunar landing, mass-optimal trajectories are obtained via numerical optimization. The upsampled trajectories are passed to a photorealistic scene generator to simulate the view of a descent camera from different heights from the surface. Combined with the optimal trajectories, the simulated frames can be used to visualise the motion field induced by the motion of the spacecraft during the descent. In the final step, an event-based camera is emulated using a video-to-event toolbox to generate synthetic event-streams from the simulated landings.

\subsection{Trajectory Generation}\label{sec:traj_gen}
Trajectory generation is based on the mass-optimal control problem described in Section \ref{sec:ocp}. By varying the descent range, velocities and attitude of the spacecraft, distinct features are captured in the image plane, simulating what a descent camera would see during braking, approach and descent to the surface. However, in the interest of dataset variability, the descent trajectories are generated from a higher starting altitude than typically observed during the terminal descent stage. Similarly, a return to zero pitch is enforced via the boundary conditions in each pinpoint landing scenario to ensure that sufficient surface features (such as craters) can be used to aid future reconstructions of pose from events.

For each landing scenario, an initial set of boundary conditions is defined as the base profile. Optimisation of the base trajectory is performed by discretising the corresponding OCP \eqref{eq:ocp_constraints} over $N$ nodes via Hermite-Simpson direct collocation; the resulting nonlinear problem is then passed to the Sparse Nonlinear OPTimizer (SNOPT)\cite{Gill2002} for optimisation. Additional optimal trajectories can then be derived from the base trajectory by sampling new conditions from preselected parameter ranges (see Table \ref{tab:landing_profiles}) and in turn applying a continuation method.
 
\begin{figure}[ht]
  \centering
  \includegraphics[width=\linewidth]{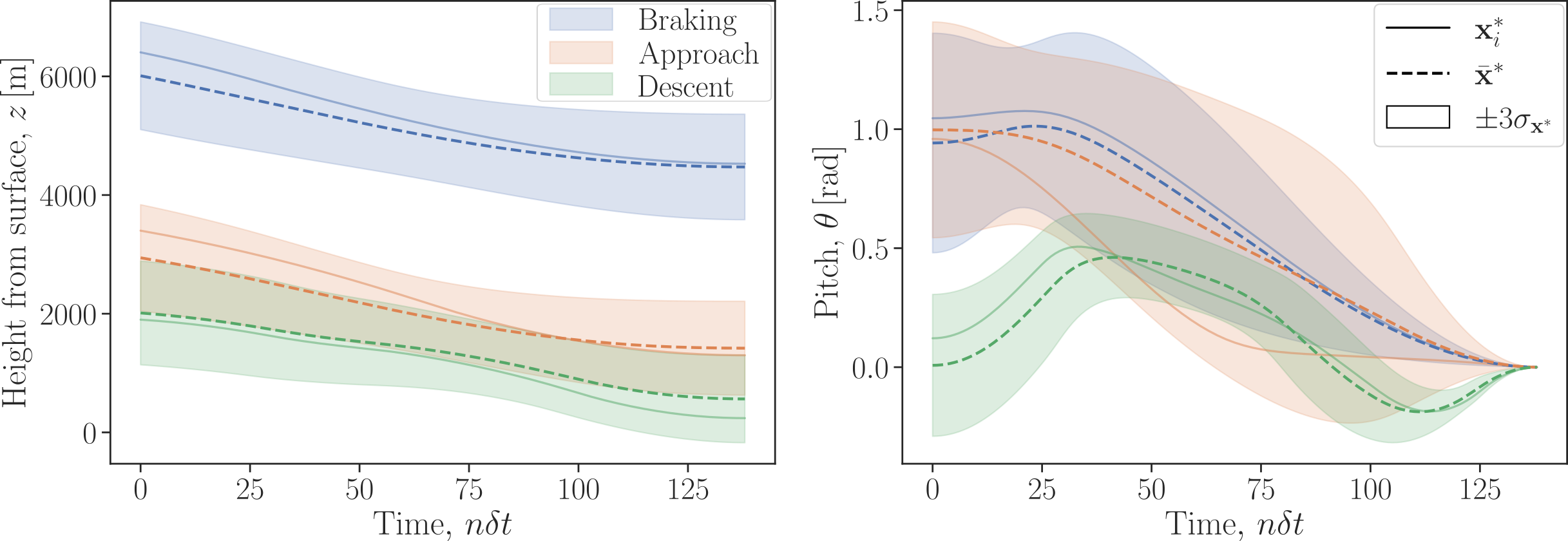}
  \caption{Height and pitch highlights from a sample dataset of 500 optimal trajectories combining braking, approach and descent characteristics (solid: sample trajectory $\mathbf{x}^*_i$; dashed: mean trajectory $\bar{\mathbf{x}^*}$; filled: $\pm 3 \sigma_{\mathbf{x}}^*$ standard deviations from the mean)}
  \label{fig:dataset_sample}
\end{figure}

 This approach is used to generate a dataset of 500 optimal trajectories $\mathbf{x}^*_i$ that showcase the different features of the pipeline. Figure \ref{fig:dataset_sample} shows the resulting distribution of height from the surface and pitch across the three landing scenarios, where the time $n \delta t$ denotes the $n$th control interval given that the timestep $\delta t = \frac{t_f}{N}$ varies with each trajectory. 

The next step in the pipeline consists in upsampling the optimal trajectories from $N$ nodes to the desired number of frames per second ($\rm{fps}$). Given the magnitude of the relative velocities of features in the scene, $\rm{fps}=100\,\rm{Hz}$ is determined to be sufficiently high a frame rate to generate smooth footage of the PANGU simulated landing for the video-to-event conversion. As a result, 1D cubic spline interpolation is applied to each of the 500 trajectories, such that $t_f * \rm{fps}$ camera viewpoints can be rendered in the scene generator.

\subsection{Frame Rendering}\label{sec:pipeline_frames}

The sequence of camera poses along the interpolated trajectories are passed to the PANGU simulator to control the camera viewpoint during the simulated descent. The camera model in PANGU is treated as a pinhole camera with camera matrix
\begin{equation}
    \mathbf{K} = \begin{bmatrix}
        f & s & c_x \\
        0 & a f & c_y \\
        0 & 0 & 1
    \end{bmatrix}
\end{equation}
where, in the idealized case, the aspect ratio $a=1$, the skew $s=0$ and the principal point $(c_x, c_y) = (W/2, H/2)$. Drawing inspiration from the characteristics of descent cameras found on previous lunar landers, we set the frame size to $(W, H) = (1024, 1024)$ such that
\begin{equation}
    f = \frac{W}{2\tan\frac{\vartheta}{2}}
\end{equation}
with the field of view set to $\vartheta = 45^{\circ}$. These characteristics allow the simulated pinhole camera to capture a wide range of features across the three landing scenarios, including the lunar horizon during high-pitch phases and the surface craters as they enter/leave the field of view. 

The target body for the simulation of the landing trajectories is a mid-resolution 3D lunar model made available by PANGU. The whole moon model uses projected digital elevation maps (DEM) to recreate the cratered lunar landscape with high-resolution features concentrated along a strip around the equator to support landing simulations. A suitable landing site is selected along this strip near the lunar south pole which, combined with low Sun elevations, recreates challenging lighting conditions that highlight the perceived benefits of dynamic vision sensing for vision-based navigation. Additionally, the simulator offers several options for the directional reflectance properties of the body, such as the Lambert and Hapke models, which allow for realistic surface shadow rendering in different lighting conditions. Table \ref{tab:landing_profiles} summarises the parameter ranges considered in each landing scenario.
\begin{table}[ht]
\begin{center}
\caption{Summary of the landing scenario parameterisation: the target gives the $(x, y, z)$ Moon-centred coordinates of a site near the south pole, and the illumination gives the Sun's (range $r_s$, azimuth $\gamma_s$, altitude $\alpha_s$)}\label{tab:landing_profiles}
\begin{tabular}{l@{\hskip 0.5in}c@{\hskip 0.25in}c@{\hskip 0.25in}c}
\toprule
\multicolumn{1}{l}{} & Braking & Approach  & Descent\\
\midrule
Descent range $[m]$ & [6500, 4000]  & [3500, 1000] & [2500, 100] \\
Initial pitch range $[\deg]$ & [45, 60]  & [45, 75]  & [-9, 9] \\
Illumination $[m, \deg, \deg]$ & (1.496\text{e}11, 105, 7) & (1.496\text{e}11, 55, 15) & (1.496\text{e}11, 155, 2) \\
\midrule
Target $[m]$ & \multicolumn{3}{c}{(-198974, 49, 1730162)} \\
\bottomrule
\end{tabular}
\vspace*{2mm}
\end{center}
\end{table}

\subsection{Motion Field Ground Truth}\label{sec:pipeline_mf}

Having generated the ground truth trajectories, the motion field induced by the descent can be computed at each interpolated pose according to \eqref{eq:mf} as an intermediary step. Given the Moon as the target body, the assumption of a spherical landing surface is used in this work (see Section \ref{sec:mf_sphere}). Despite only being an ideal representation of the motion of 3D points projected into the image plane, the motion field finds value in its relation to the optical flow. Unlike the motion field which is computed from idealised equations of motion, the optical flow can be estimated from measurements of surface features and landmarks, which become apparent in an event-based representation (see Figure \ref{fig:events_mosaic}). By treating the optical flow as an approximation of the ground truth motion field (the validity of which is application dependent), it is possible to reconstruct the spacecraft motion by inverting \eqref{eq:mf}. The resulting inverse problem, while overdetermined, can be solved for the inverse depth map $h(x_s, y_s)$ and translational velocity $\mathbf{v}_c^{vi}$ (up to an arbitrary scaling factor) and the rotational velocity $\boldsymbol{\omega}_{vi}$ \cite{Soatto1997}.

\subsection{Synthetic Event Stream Generation}\label{sec:pipeline_events}

The last step in the pipeline converts the sequences of frames rendered by PANGU (and concatenated into videos at $\rm{fps} = 100\,\rm{Hz}$) into streams of events, thereby emulating the output of an event-based camera. The video-to-event toolbox is configured using mid-range parameter values that suit the simulated low-light conditions \cite{Hu2021}. The tuning parameters used in this work are summarised in Table \ref{tab:v2e_params}. The parameterisation results in a noisy sensor emulation where, in addition to events generated by the apparent motion of surface features, pixels fire events according to known noise models (see Section \ref{sec:event_cam}). While more realistic, this will render any reconstruction task from events more challenging.
\begin{table}[ht]
\begin{center}
\caption{Parameterisation of the emulated event-camera model}\label{tab:v2e_params}
\begin{tabular}{@{}lcc@{}}
\toprule
Parameter & Value & Unit \\
\midrule
Contrast threshold mean ($\Theta_{\rm{ON}}$ / $\Theta_{\rm{OFF}}$) & $\pm 0.25$ & N/A\\
Contrast threshold standard deviation ($\sigma_{\Theta}$) & 0.03 & N/A\\
Shot noise rate ($f_{\rm SN}$) & 5 & $\rm{Hz}$\\
Leak rate ($f_{\rm LR}$) & 0.1 & $\rm{Hz}$\\
IIR low-pass cutoff frequency ($f_{\rm LP}$) & 3 & $\rm{Hz}$\\
\bottomrule
\end{tabular}
\end{center}
\end{table}

\section{Simulation and Results}\label{sec:results}

\begin{figure}[H]
  \centering
  \includegraphics[width=\linewidth]{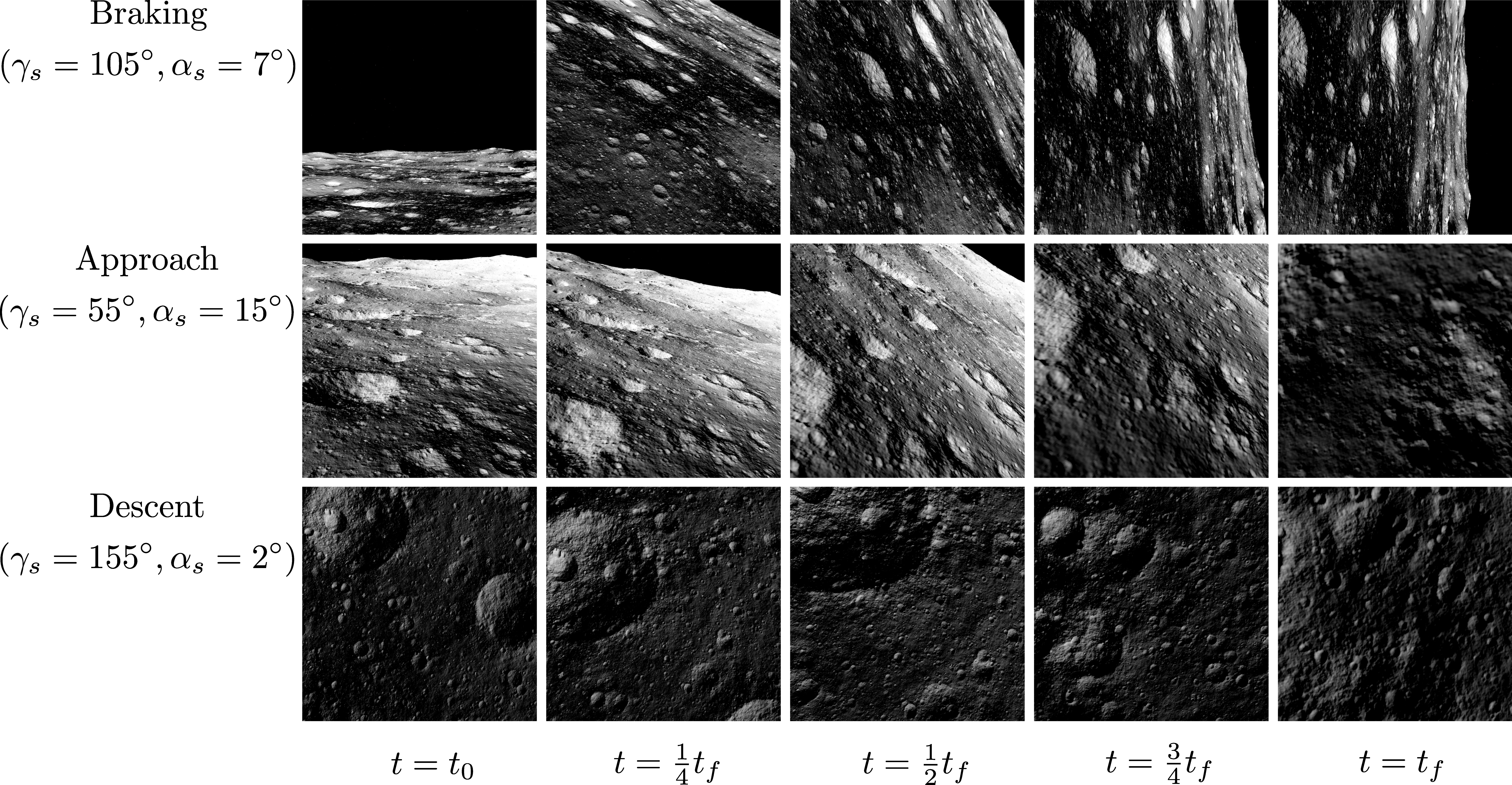}
  \caption{PANGU renders of the lunar surface at different stages of the simulated descent and under the low-light conditions given in the left margin (Hapke reflectance)}
  \label{fig:frame_mosaic}
\end{figure}

\begin{figure}[H]
  \centering
  \includegraphics[width=\linewidth]{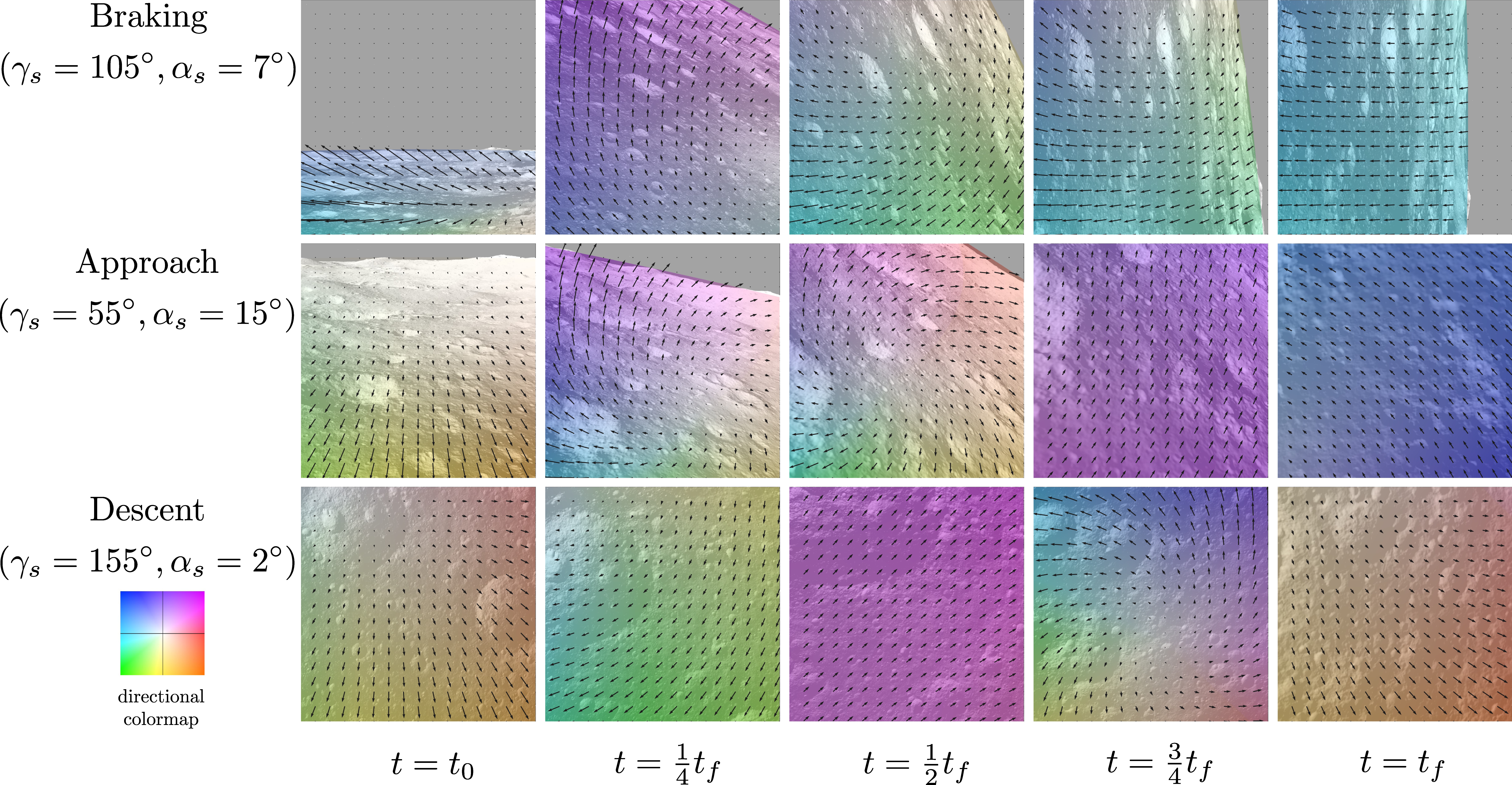}
  \caption{Ego-motion field induced by the relative motion of the camera at different stages of the simulated descent (assuming a spherical landing surface with $R=1737400 + \rm{mean}(h_{\rm{DEM}})\,\rm{[m]}$)}
  \label{fig:mf_mosaic}
\end{figure}

\begin{figure}[H]
  \centering
  \includegraphics[width=\linewidth]{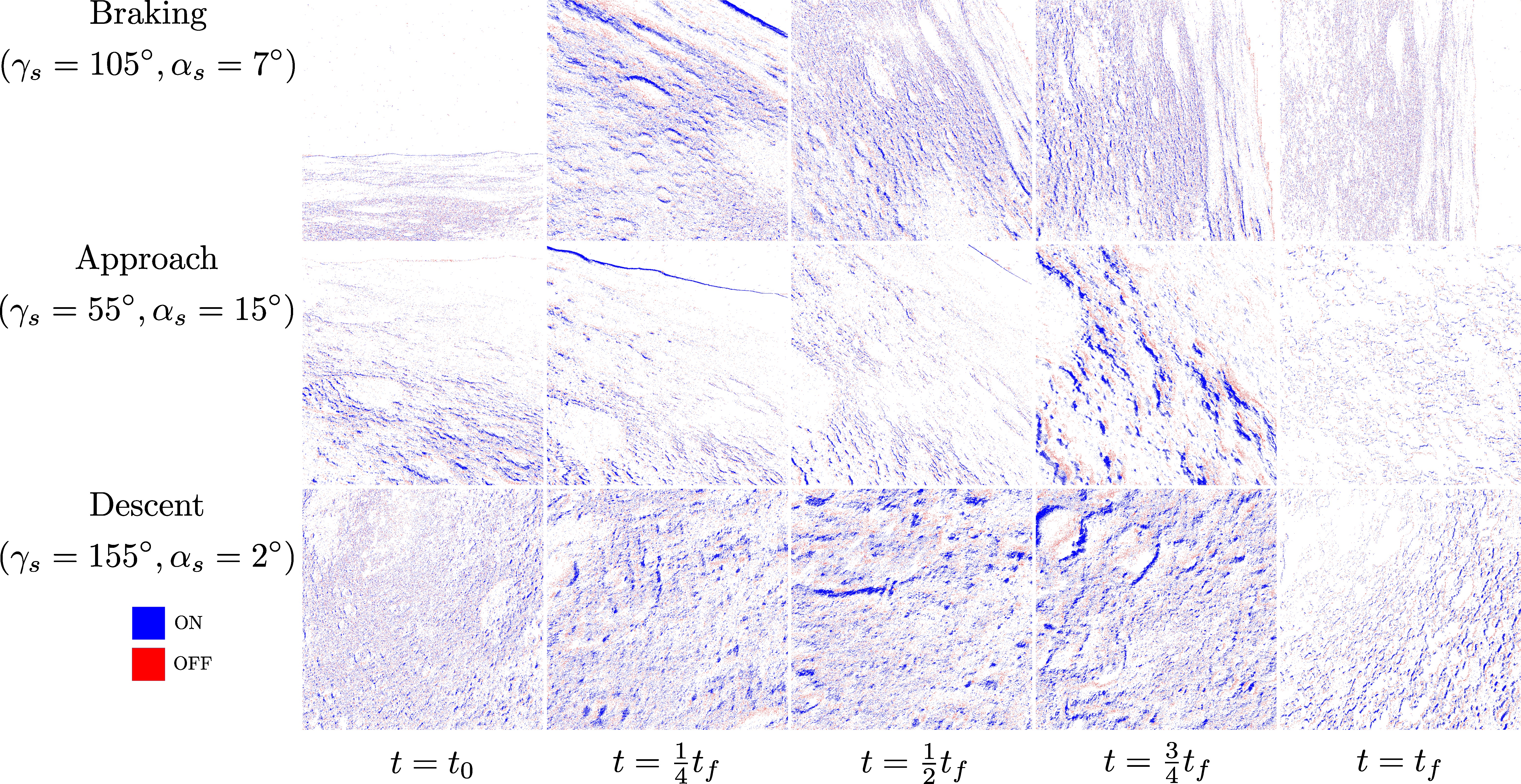}
  \caption{Event frame representation of the lunar surface at different stages of the simulated descent and under the low-light conditions given in the left margin: blue (red) pixels denote positive/ON (negative/OFF) changes in scene brightness}
  \label{fig:events_mosaic}
\end{figure}

The first output of the pipeline consists in photorealistic renders of the lunar surface generated by PANGU. Figure \ref{fig:frame_mosaic} depicts 15 scenes taken from the simulation of the three landing scenarios whose parameter ranges are described in Table \ref{tab:landing_profiles} at timesteps $t = \{t_0, t_f/4, t_f/2, 3t_f/4, t_f\}$. In both the braking and approach simulations, the spacecraft pitch decreases over the entire trajectory, with significant roll and yaw motion. The last row describes a near-ventral descent to the surface.
The second pipeline output is related to the motion field, discussed in Section \ref{sec:cam_eom}. Figure \ref{fig:mf_mosaic} overlays the motion field computed from the trajectory ground truth onto the renders of Figure \ref{fig:frame_mosaic}. While the motion field is computed at each pixel in the image, a sparser quiver plot is shown for clarity. Alternatively, a directional flow colour code is used to map the motion field to RGB values, creating smooth colour gradients over the full frame \cite{Baker2007}. By including motion ground truth data in this format, it is envisioned that the pipeline can support frame-based motion reconstruction methods, such as those based on convolutional neural networks.
By design, the pipeline outputs event streams in a spatiotemporal format. However, in order to have a meaningful comparison between the original frames and the spatial distribution of synthetic events around a small temporal neighbourhood, events can be accumulated into \textit{event frames}. The event frames shown in Figure \ref{fig:events_mosaic} accumulate the events over a short time window of $\Delta t = 10\,\rm{ms}$ centred on each of the timesteps $\{t_0, t_f/4, t_f/2, 3t_f/4, t_f\}$.
An event-based representation of the lunar horizon can clearly be seen in the top-left event-frame of Figure \ref{fig:events_mosaic}. As a result of the noisy event-based camera emulation, significant noise can be observed beyond the horizon line where, except for a few stars, there exist no visual features that would trigger events. In general, salient features in the scene such as craters, boulders and shadows are highlighted with events as a result of the motion of the camera. Given longer exposure times, this will lead to near saturation of the middle event frames ($t_f/4, t_f/2, 3t_f/4$), when the motion of the camera is most pronounced, compared to fewer events in the final time $t_f$, when the scene is nearly static. These results highlight the motion-dependent properties of the output of an event-based camera.

% TODO: consider adding incremental event frame mosaic
% \begin{figure}[ht]
%   \centering
%   \includegraphics[width=\linewidth]{figures/events_mosaic.jpg}
%   \caption{Event frame representation of the lunar surface at different stages of the simulated descent and under the low-light conditions given in the left margin: blue (red) pixels denote positive/ON (negative/OFF) changes in scene brightness}
%   \label{fig:events_mosaic}
% \end{figure}

\section{Discussion}\label{sec:discussion}

With the current means available, it is difficult to determine the extent to which the synthetic events approximate what an event-based camera would see upon approach of the lunar surface. Nonetheless, the proposed event-based dataset provides a representation of surface features over time that is fundamentally different from the frame-based state of the art, and allows practitioners to develop novel tools to extract navigation information from events. For this reason, the accumulation of events into frames, as done above in the interest of qualitative analysis, is often found to be impractical. While Figure \ref{fig:events_mosaic} accumulates events over a short time window into a single frame, there exist alternative methods based on event counts or the coverage of a fixed image area; the difficulty lies in determining which method best captures the features relevant to the application under study. In doing so, however, one ends up emulating the operation of a frame sensor, which does not take full advantage of the asynchronous and sparse nature of events.

Given the simplified visual navigation geometry (i.e., the coincident camera and spacecraft coordinate frames) and the use of an idealised camera model, the pipeline proposed in this work serves as a baseline for more realistic event-based dataset generation from landing trajectories in the future. An important limitation of the current setup is the size and variability of the generated dataset. The formulation of the landing problem as a minimum-mass OCP ensures that the generated optimal trajectories remain smooth. This prevents the occurrence of any kind of jerky camera motion or correction manoeuvres which would translate into very distinct event distributions. Moreover, the trajectory generation is currently based on the variation of only a subset of the boundary conditions. Future work may look at expanding the parameter space beyond the ranges proposed in Table \ref{tab:landing_profiles} in an attempt to improve the resulting distributions of landing profiles. The pipeline is designed to be modular in light of these anticipated additions, and it is straightforward to replace the trajectory optimisation component with a more suitable source of landing trajectories. As such, the pipeline may be adapted to take in camera poses from an external software package (e.g., SPICE kernels).

Dataset heterogeneity can also be achieved by varying the simulated landing environments. While the generation of realistic shadow maps in PANGU can prove computationally expensive, it is a readily available means to alter the distribution of events generated for a given trajectory. Given space constraints, this work associates each landing scenario with a different set of low-light conditions. Repeating the landing simulation with different Sun positions could effectively augment the dataset as the changing shadows would lead to different event streams. This would provide additional material for a comparative analysis of event distributions.

Event-based datasets for navigation and landing can be prepared so as to replicate space environments as faithfully as possible. While low-light conditions are discussed to some extent in this study, additional artefacts such as those induced by high-energy particles are not. Radiation effects are already supported by scene generators like PANGU, and it may be worthwhile to study their impact on the generation of event streams in a follow-up study.

A limitation of the generation and use of ground truth motion field data for navigation and landing is the assumption of \textit{a priori} knowledge about the target body. While the assumption of a planar or spherical surface can be justified in considering landings on large spherical bodies such as the Moon, this is no longer the case for target bodies with irregular shapes such as asteroids. While a more detailed DEM may be used in these cases, such information may not always be available. Despite using the spherical assumption to overlay the motion field onto the frames in Figure \ref{fig:mf_mosaic}, there still remains an offset owing to the use of a DEM on top of a perfect sphere in PANGU. A better understanding of these models is necessary in order to obtain a reliable estimate of the ground truth motion field.   

\section{Conclusion}

This work proposes a data pipeline for generating synthetic event-based vision datasets from the point of view of an event-based camera for navigation and landing applications. The pipeline's features are showcased in this work through the generation of an event-based dataset for simulated lunar landings. First, a minimum-mass optimal control problem is solved based on trajectory specifications that approximate the braking, approach and descent stages of a typical lunar landing. Each optimal trajectory is then used as a baseline to solve additional landing profiles, thus augmenting the dataset. Knowing the pose of the camera at each point along the trajectories, a video of the lunar landing can be generated using the PANGU scene generator for planetary bodies and asteroids, where the lighting conditions can be varied to further contribute to dataset variability. In the final step, the videos are passed through a video-to-event converter which emulates the event-camera and outputs spatiotemporal streams of asynchronous events. By aggregating events over short time windows, an event-based representation of the lunar surface at different stages of the landing can be obtained and compared to the original rendered scenes and ground truth data. Additional information, such as the motion field derived from the ground truth trajectories, is made available as part of the pipeline in anticipation of reconstruction methods based on optical flow.

In follow-up work, we will further extend the dataset by considering additional landing scenarios and low-light conditions. An upcoming open-data challenge will be based on the resulting landing dataset to engage the wider computer vision communities in determining how best to process event-based data for navigation and landing\footnote{The open data challenge will be hosted on the Kelvins data competition platform hosted by the European Space Agency's Advanced Concepts Team: \url{https://kelvins.esa.int/}}. The success of previous similar initiatives on spacecraft pose estimation \cite{Kisantal2020}\cite{Park2022} testifies to the effectiveness of competitive data challenges in garnering interest and solutions to novel optical navigation opportunities. Beyond the data challenge, the event-based landing dataset is envisioned to support investigations into future onboard opportunities for event- and vision-based navigation, including motion estimation based on optical flow, surface feature identification and tracking, and terrain relative navigation. By releasing this pipeline, we hope to promote the creation of new datasets for event-based navigation around other planetary bodies and asteroids, and to support the development of state of the art event processing tools for future space missions.

% Bibliography
\pagebreak
\printbibliography
\end{document}